\documentclass[10pt,twocolumn,letterpaper]{article}

\usepackage{cvpr}              

\usepackage{graphicx}
\usepackage{amsmath}
\usepackage{amssymb}
\usepackage{booktabs}
\usepackage{cite}
\usepackage{makecell}
\usepackage{array}
\usepackage[pagebackref,breaklinks,colorlinks]{hyperref}

\usepackage[capitalize]{cleveref}
\crefname{section}{Sec.}{Secs.}
\Crefname{section}{Section}{Sections}
\Crefname{table}{Table}{Tables}
\crefname{table}{Tab.}{Tabs.}

\begin{document}


\title{Weakly-supervised Generative Adversarial Networks for medical image classification}

\author{Jiawei Mao$^{1}$   \quad Xuesong Yin$^{1*}$ \quad Yuanqi Chang$^{1}$ \quad Qi Huang$^{2}$  \\ 
$^{1}$School of Media and Design, Hangzhou Dianzi University, China \qquad \\
$^{2}$School of Biological \& Chemical Engineering, Zhejiang University of Science and Technology, China\qquad \\
{\tt\small\{jiaweima0,yinxs,211330020\}@hdu.edu.cn }\\
{\tt\small\{qihuang\}@zust.edu.cn }\\
}
\maketitle

\begin{abstract}
Weakly-supervised learning has become a popular technology in recent years. In this paper, we propose a novel medical image classification algorithm, 
called Weakly-Supervised Generative Adversarial Networks (WSGAN), which only uses a small number of real images without labels to generate fake images or mask images to enlarge the sample size of the training set. 
First, we combine with MixMatch to generate pseudo labels for the fake images and unlabeled images to do the classification. Second, contrastive 
learning and self-attention mechanism are introduced into the proposed problem to enhance the classification accuracy. Third, the problem of mode collapse is well addressed by cyclic consistency loss. 
Finally, we design global and local classifiers to complement each other with the key information needed for classification. 
The experimental results on four medical image datasets show that WSGAN can obtain relatively high learning performance by using few labeled and unlabeled data. 
For example, the classification accuracy of WSGAN is 11\% higher than that of the second-ranked MIXMATCH with 100 labeled images and 1000 unlabeled images on the OCT dataset. 
In addition, we also conduct ablation experiments to verify the effectiveness of our algorithm.
   
\end{abstract}

\section{Introduction}
\label{sec:intro}
Due to the limitation of conditions in the field of Medicine, one always confront those issues: 
(1) It is difficult to obtain medical images because of the protection of personal privacy and the high cost of medical images; 
(2) Since each medical image needs to be carefully diagnosed by experts, it is expensive to labeling medical images; 
(3) Compared to other natural images, most medical images are gray images with relatively simple structure. 
Moreover, many lesion areas occupy only a small part of the whole image so that capturing the feature of these images becomes troublesome. 
Thus, classifying such medical images has become a challenging task. 
Weakly-supervised classification has been successfully applied to scenarios with a small number of data and class labels and thus attracted more and more attention in recent years \cite{chawla2005learning,li2019towards,zhou2018brief}.  

\begin{figure}[h]
	\centering
	 \includegraphics[width=1\linewidth]{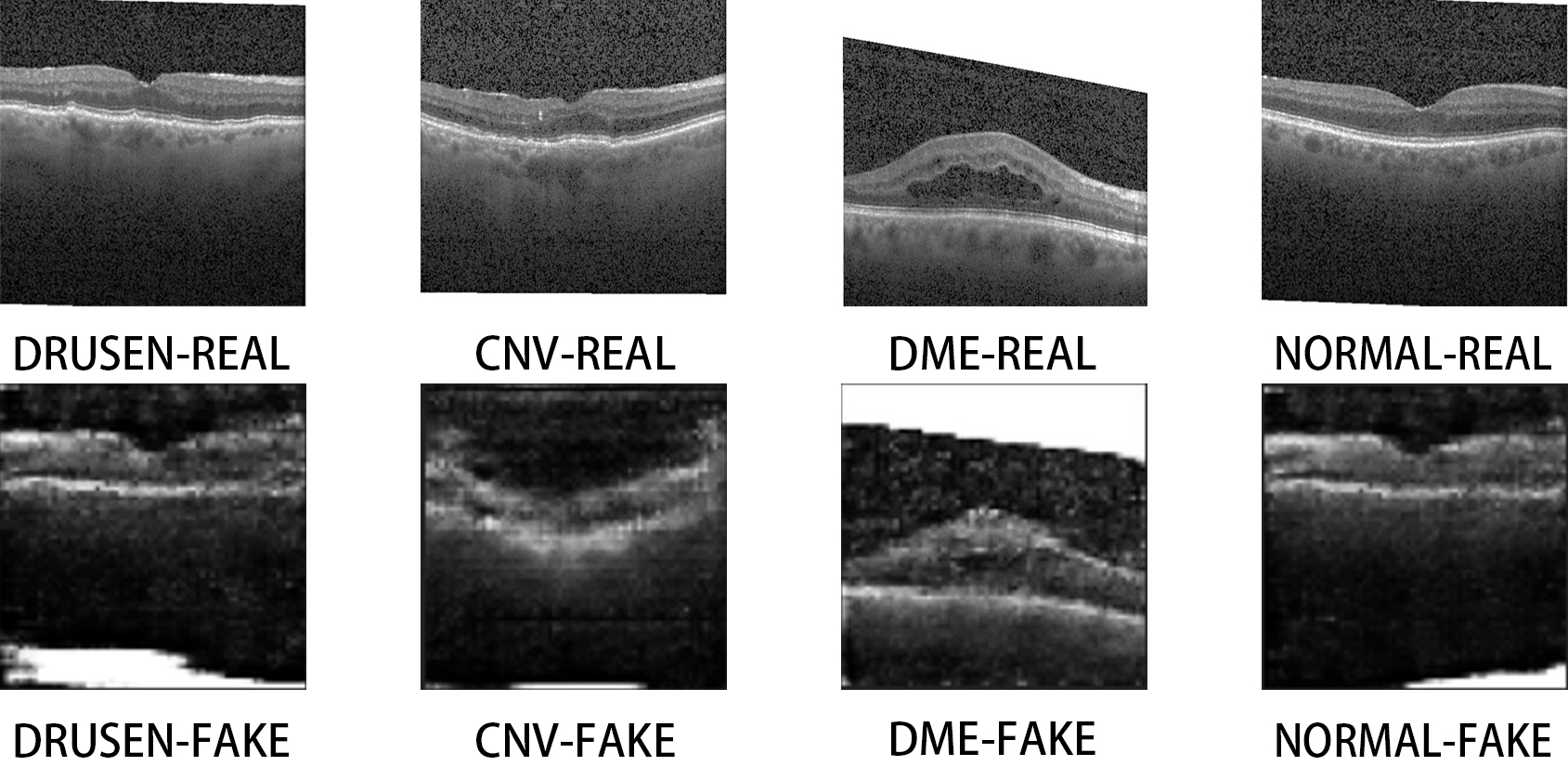}
	 \caption{Optical coherence tomography image generation with only 1000 real OCT images. The first row is real images and the second row is mask images.}
  \end{figure}

With the rapid development of the Convolutional Neural Network (CNN)\cite{krizhevsky2012imagenet,lecun2015deep}, experiments on medical image datasets \cite{bhowal2021choquet,javaheri2020covidctnet,kamran2019optic,kamran2020improving,krishnan2021vision,wang2020weakly,pinckaers2021detection} have shown that many algorithms have made great progress in medical image classification, 
which addresses the above problems to some extent.

Pseudo-labeling classification\cite{lee2013pseudo} can be used to label the medical image. For each unlabeled image, it can generate pseudo tags to enhance image classification by respecting the feature of each image. 
The feature can be learned effectively by considering the common feature of two images with different data augment. 
Generally, we often face the problem of both insufficiency images and labels in real-world medical applications. 
Despite pseudo-labeling classification being used to solve the problem of image annotation, it is ineffective to deal with such an issue due to the lack of unlabeled images. 
Pseudo-labeling assigns the machine-learned labels to unlabeled images, thus greatly facilitating the development of incomplete supervision in weakly-supervised learning.

Recently, Generative Adversarial Networks (GANs)\cite{goodfellow2014generative} can successfully generate images that confuse the real with the false. 
It is used in a variety of applications such as image generation\cite{brock2018large,karras2019style,karras2020analyzing,zhang2019self}, style conversion\cite{chen2018cartoongan,isola2017image,wang2020learning,zhu2017unpaired} , conditional image generation\cite{dash2017tac,mirza2014conditional,odena2017conditional}, 3D modeling\cite{wu2016learning,zhu2018visual} , semantic segmentation\cite{luc2016semantic}, 
image restoration and super-resolution\cite{ledig2017photo,wang2018esrgan}. Similarly, GAN can also be used in the field of image classification, which can expand the number of training images by generating fake images. 
However, the main problem faced by the application of GANs in classification tasks is the lack of labels in generated images. 

Motivated by recent progress in adversarial learning and weakly-supervised learning, in this paper we propose a weakly-supervised classification generative adversarial network (WSGAN), 
WSGAN can obtain a good generator by adversarial learning based on the FID metric, and then we fix the generator to fine-tune our classifier. 
Finally, we provide pseudo labels for the generated images to train our classifier jointly with the labeled real images. 
We achieve good classification results on the medical datasets with sparse images. 
Consequently, the proposed WSGAN solves the above problems of insufficient unlabeled data in weakly-supervised medical classification and the lack of labels in generated images. 

The highlights of our approach are the following four points:

(1)	We strengthen the power to solve the problem of model collapse in contrastive learning by cyclic consistency loss. 
In the pre-training process of contrastive learning, our algorithm prevents all images from being encoded into the same vector by adding a decoder to re-decode the middle vector back to the input image.

(2)	Aiming at the poor classification performance of GANs, we exploit the trained generator to produce fake images which are used in classification with the real images. 
We verified the correctness of our algorithm by conducting comparative experiments on SSGAN.

(3)	To the best of our knowledge, our approach is the first to combine pseudo-labeling classification with generative adversarial networks. 
We increase the number of unlabeled images by generating images and exploiting pseudo labels to provide class labels for the generated images. 
Our approach can achieve very good results with a small number of medical images.

(4)	Richer information is captured by our classifier, which adopts global and local architectures. 
We add a spatial self-attention mechanism and a pixel self-attention mechanism to each classifier to focus on the most relevant locations of the global image, thus improving the model's odds to find lesion regions.

\section{Related Works}
\subsection{Pseudo-labeling}
Currently, weakly-supervised image classification mainly uses pseudo-labeling to create a pseudo label for each unlabeled image. 
For example, MixMatch\cite{berthelot2019mixmatch} generates pseudo tags on unlabeled images through twice data enhancements and allows these images to get consistent classification results by the classifier. 
Finally, pseudo tags and real tags are mixed up for weakly-supervised classification. Hieu Pham et al.\cite{pham2021meta} proposed the Meta Pseudo label method, which is divided into teacher network and student network. 
The teacher network generates false labels on unlabeled images. The student network is then trained by combining these pseudo-labeled images with labeled images. 
A new robust pseudo label selection framework is designed to greatly reduce the influence of poor network calibration on the pseudo-label process by Rizve et al.\cite{nayeem2021defense}. 
Our method mainly uses GANs to increase the amount of unlabeled medical images, so that there are enough images to manufacture pseudo labels.

\subsection{Generative Adversarial Networks}
Until recently, GANs have made great progress in image synthesis. 
SAGAN\cite{zhang2019self} and BigGAN\cite{brock2018large} which are equipped with a non-local module self-attention mechanism\cite{vaswani2017attention,wang2018non} can generate realistic images through dependency between pixels in each layer. 
In Improved Techniques for Training GANs\cite{salimans2016improved} Goodfellow et al. come up with the idea about semi-supervised learning Generative adversarial network, which opens a door in GANs classification. 
On CatGAN\cite{2015Unsupervised}, Jost Tobias Springenberg helps the classifier training by learning the labeled and unlabeled samples from the data distribution. 
It obtains the representation information in the shared structure, using only a small number of images. 
However, the result of GAN classification is not ideal, Dai et al.\cite{dai2017good} explained why and suggested that a bad generator was needed to gain a good classification result. 
A good generator is obtained by our algorithm and then we use the fixed generator to output the mask image for classification. 
In addition, with the emergence of GANs such as SinGAN\cite{shaham2019singan}, it is possible to use fewer real images to get high fine-grained images.

\begin{figure*}[h]
	\centering
	 \includegraphics[width=1\linewidth]{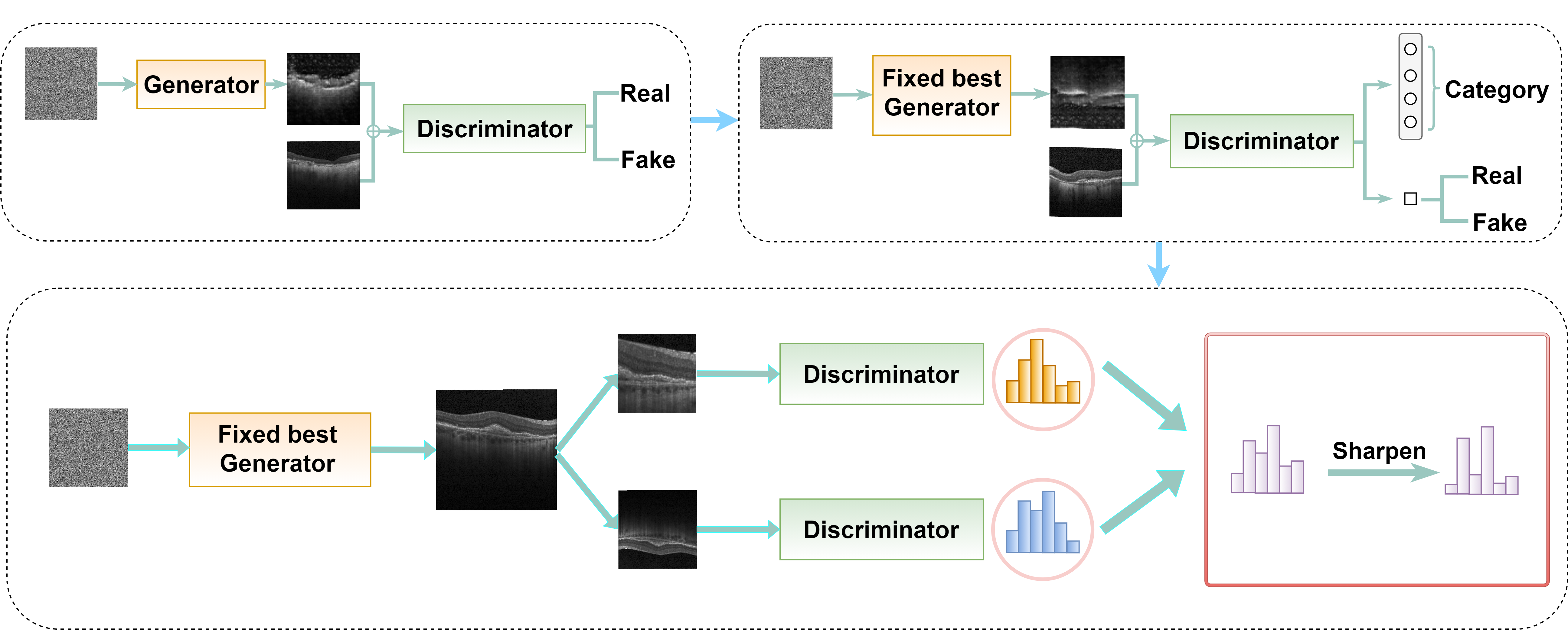}
	 \caption{An overall pipeline of the proposed method.}
  \end{figure*}

\subsection{Contrastive Learning}
The main idea of contrastive learning is to draw the mapped latent vectors from the same image closer and push latent vectors from different images farther. 
The SIMCLR\cite{chen2020simple} network proposed pre-trained by contrastive learning has great success in semi-supervised classification, which gained 92.6\% top1-acc on ImageNet-10\% labeled images. 
Yet, contrastive learning often has the issue of mode collapse, that model maps all inputs into the same constant vector. 
To solve this problem, different solutions are proposed. Jing et al.\cite{jing2021understanding} proposed a new contrastive learning method called DirectCLR, which directly optimizes the encoder (i.e. representation space) without relying on the trainable projector. 
Experiments show that this algorithm can effectively prevent dimension collapse and mode collapse. Chen et al\cite{chen2021exploring}. use stopping the gradient to solve the problem of the mode collapse in contrastive learning, 
which also attained better results. Inspired by Cycle GAN’s\cite{zhu2017unpaired} cycle consistency, we believe that if the decoder is used to decode the vector back to the input image, the image cant be encoded into the same vector.
 We also performed an ablation experiment to prove the validity of our decoder.

\subsection{Weakly-supervised learning (WSL)}
WSL is divided into three main categories: incomplete supervision, inexact supervision, and inaccurate supervision. 
Incomplete supervision\cite{dehghani2017neural,he2021online} means that only a small portion of the training images is labeled, while most images are unlabeled. It is not enough to train a good model with these labeled images. 
Inexact supervision\cite{bouveyron2009robust} is a situation where we have some supervised information, but it is not as precise as we would like it to be. A typical case is when we only have coarse-grained labeling information. 
Inaccurate supervision\cite{dubost2020let,shu2020detecting,wang2014label} is concerned with situations where the supervised information is not always true, in other words, some labeling information may be wrong. 
We only use a small number of labeled and unlabeled images to construct a predictive model. Therefore, our algorithm belongs to the incomplete classification.

\section{Proposed Method}
Contrastive learning can reduce the distance between samples of positive cases and enlarge the distance between samples of negative cases in latent space. 
Therefore, we introduce contrastive learning into our model. Our training process is mainly divided into four steps. In step 1, we pre-train the feature layer of ResNet50 to extract key image features. 
In order to provide sufficient unlabeled images for weakly-supervised classification, we use the GAN to simulate the labeled image’s distribution with random Gaussian noise in step 2. 
In step 3, we first let the model to learn certain features of the real image and fake image. In step 4: according to the method of MixMatch, a pseudo label is created for the generated image in classification.

\subsection{Contrastive Learning Pre-training With Decoder}
\begin{figure}[h]
	\centering
	 \includegraphics[width=1\linewidth]{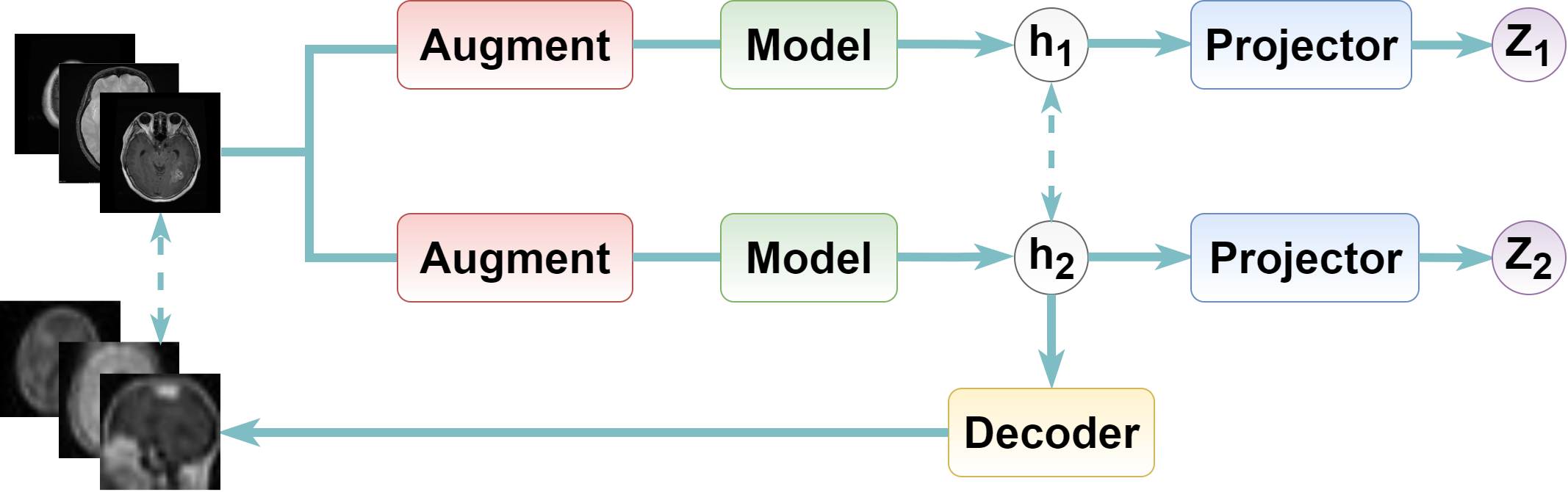}
	 \caption{A framework for contrastive learning with a decoder.} 
  \end{figure}
WSGAN first perform data enhancement operations such as random cropping, random color distortion, and flipping on medical images. 
Then a batch of 2N data-enhanced images are input into the pre-training network together to obtain the latent vector H. Next, we input H into the projection layer to obtain the final vector Z. 
In the end, by minimizing the infoNCE loss, we improve the similarity of the same group images and reduce the similarity of different groups images, thus helping our model to find the lesion area better. 
Self-supervised learning generally faces the problem of mode collapse. To solve this issue, we use the convolution part of the Resnet50 network as an encoder and design a new decoder to reconstruct the hidden vector H back to the image. 
By calculating the Mean Square Error loss (MSE) \cref{eq:2}, we make the input image and the reconstructed image as similar as possible.

\begin{figure}[h]
	\centering
	 \includegraphics[width=1\linewidth]{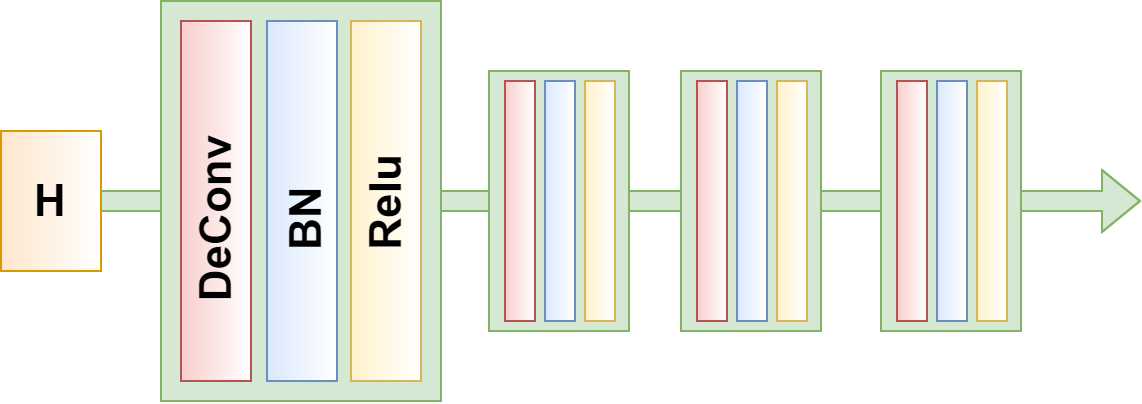}
	 \caption{A framework for our decoder model. All convolution transpose layers inside the decoder have 4 × 4 kernels.} 
  \end{figure}

$S_{i,j}$ is used to represent the distance between the final vector $Z_{i}$ and $Z_{j}$
\begin{equation}
	\begin{split}
	&s_{i,j} =\frac{Z_{i}^{}Z_{j}}{\left \| Z_{i }  \right \|\left \| Z_{j}  \right \|  }
	\end{split}
  \label{eq:1}
  \end{equation}

The positive pair of the infoNCE loss (L(i,j)) is shown: 
\begin{equation}
	\begin{split}
		{{l}_{(i,j)}}=-\log \frac{\exp ({{s}_{i,j}}/T)}{\sum\limits_{k=1}^{2N}{{{1}_{[k\ne i]}}}\exp ({{s}_{i,k}}/T)}
	\end{split}
  \label{eq:2}
  \end{equation}F

Our pre-training losses are shown as follows.
\begin{equation}
	\begin{split}
	&loss=\frac{1}{2N}\sum\limits_{m=1}^{N}[l(2m-1,2m)+l(2m,2m-1)] \\ 
	&+\frac{1}{N}\sum\limits_{m=1}^{N}(X_m-D_{ec}(H_m))^2
	\end{split}
  \label{eq:3}
  \end{equation}

\subsection{Image Generation and Semi-supervised Learning}
Since GAN can be used for classification, many algorithms extend it to feature learning. Different from the MMGAN [13] , the discriminator of the semi-supervised classification GAN has n+1 outputs. 
The first n categories are responsible for predicting specific categories, and the last one (n+1) is responsible for predicting true or false. 
Despite GAN can be used to classify those images, the image generated by GAN is unclear and the image classification accuracy is not high when the final training converges. 

The loss function of the semi-supervised classification GAN is formulated as
\begin{equation}
	\begin{split}
	\mathcal{L}_S=-\mathbb{E}_{x,y\sim{P}_{data}(x,y)}{\log }_{{p}_{\bmod el}}(y|x,y<n+1)] \\ 
	\end{split}
  \label{eq:4}
  \end{equation}
  \begin{equation}
	\begin{split}
   &\mathcal{L}_G=-\{\mathbb{E}_{x\sim{p}_{data}(x)}\log [1-{p}_{\bmod el}(y=n+1|x)] \\ 
   &+\mathbb{E}_{x\sim~Generator}{\log }_{{p}_{\bmod el}}(y=n+1|x)\}
	\end{split}
  \label{eq:5}
  \end{equation}

The objective functions (3) and (4) in SSGAN force its model to generate high-definition images and classify the input images at the same time. 
It incurs such a chosen issue, which balances between the accuracy of image classification and the quality of image generation. Therefore, when the classifier works well, the generated images are not clear\cite{salimans2016improved}.

To address the above issue, we separate the GAN training process from the image classification. WSGAN uses the GAN objective function to train a good generator. 
With such a generator, we generate images for classification. The image is still divided into n+1 categories. 
The first n categories are used to classify those images; the last one is used to distinguish true or false by using the form of least squares \cite{mao2017least} \cref{eq:5}. We exploit the cross-entropy loss to classify those images. 

Let x and z denote the image and the random noise, respectively. The loss function in step 2 for training the generator is defined as follows.
\begin{equation}
	\begin{split}
	&L=\mathbb{E}_{x\sim{p}_{x}}{(D(x)-b)}^{2}+{\mathbb{E}_{z\sim{p}_{z}}}{(D(G(z))-a)}^{2} \\ 
	&L=\mathbb{E}_{z\sim{p}_{z}}{(D(G(z))-c)}^{2}
  \end{split}
	\label{eq:6}
	\end{equation}
	\begin{figure}[h]
		\centering
		 \includegraphics[width=1\linewidth]{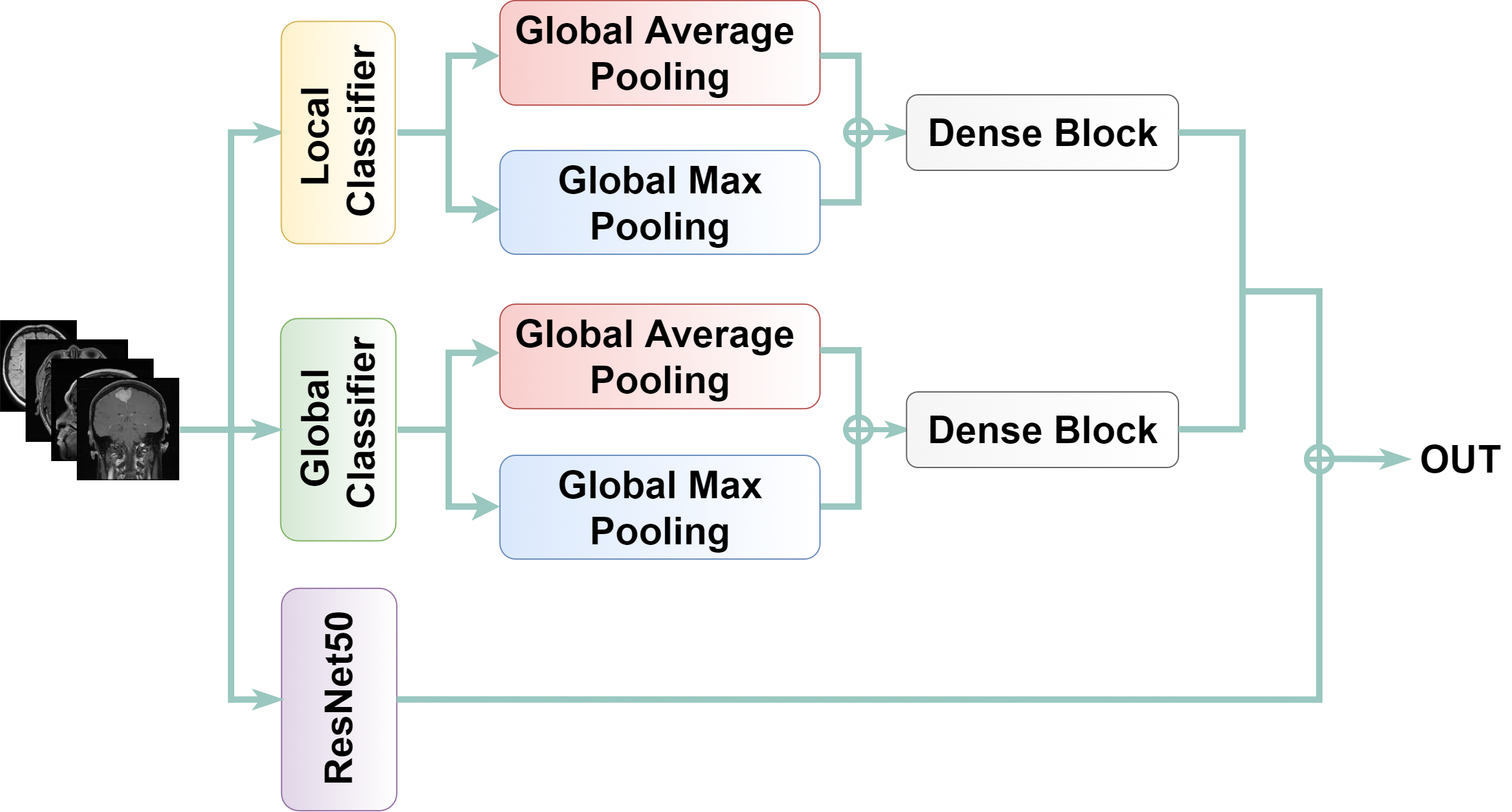}
		 \caption{Our total classifier architecture. The model we proposed consists of a local classifier and a global classifier, which both use residual architecture and attention mechanisms.}
	  \end{figure}

There are two classifiers in our model: a global classifier and a local classifier. 
Our basic module uses a spectrally normalized convolution kernel for residual linking so that the training process of our model is steady while retaining sufficient information.
\begin{figure}[h]
	\centering
	 \includegraphics[width=0.7\linewidth]{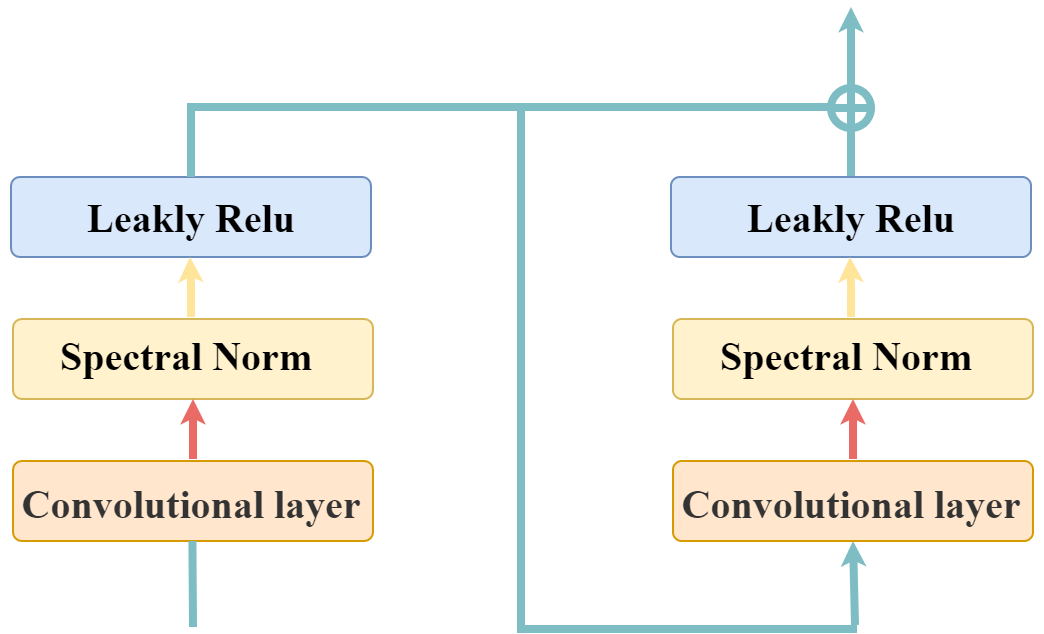}
	 \caption{Our convolutional block in the classifier.}
  \end{figure}
In both classifiers of our model, two convolutional blocks (CB) for down-sampling are used to extract feature information. 
WSGAN employs spatial self-attention and pixel self-attention mechanisms which work together to find the relevant features among the extracted features. 
Subsequently, we sum the features computed by these attention mechanisms and continue to perform convolutional down-sampling. 
The difference between global and local classifiers is that the former uses only two layers of CB, the latter utilizes five layers of CB. 
Thus, the global classifier performs deeper feature compression on the input image. By enlarging the network depth, the global classifier has a greater receptive field. 
\begin{figure}[h]
	\centering
	 \includegraphics[width=1\linewidth]{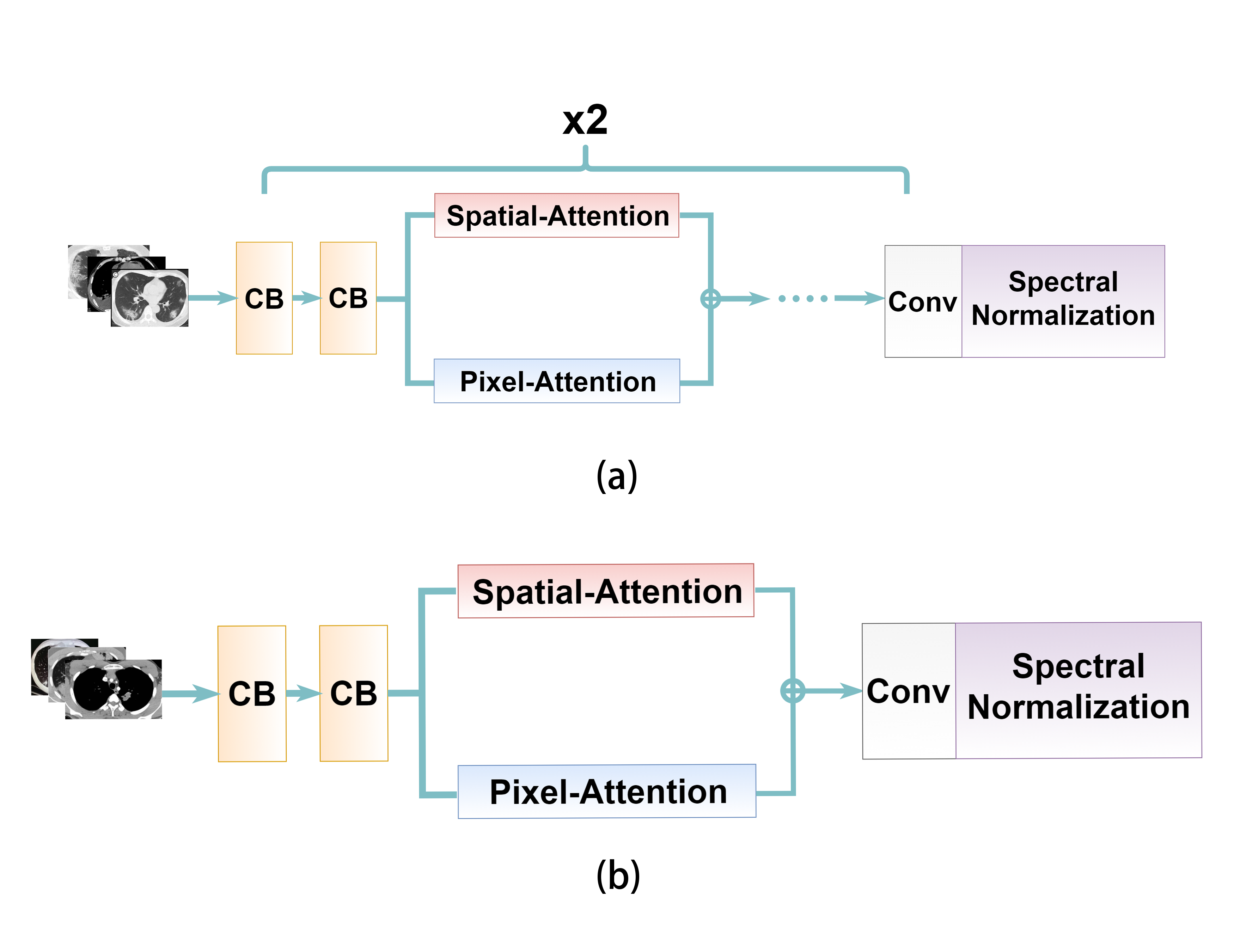}
	 \caption{ Our local classifier and global classifier: (a) is the global classifier, and (b) is the local classifier.}
  \end{figure}
Since our discriminator is very powerful, real and fake images are immediately identified when the discriminator is trained to a certain level. 
At this time, any image output by the generator is not useful. Thus good generator often fails to be obtained after training for too long. 
For the purpose of achieving a better generator, we introduce the FID\cite{heusel2017gans} evaluation index to evaluate the quality of each generator during the training process.
\begin{figure}[h]
    \centering
     \includegraphics[width=1\linewidth]{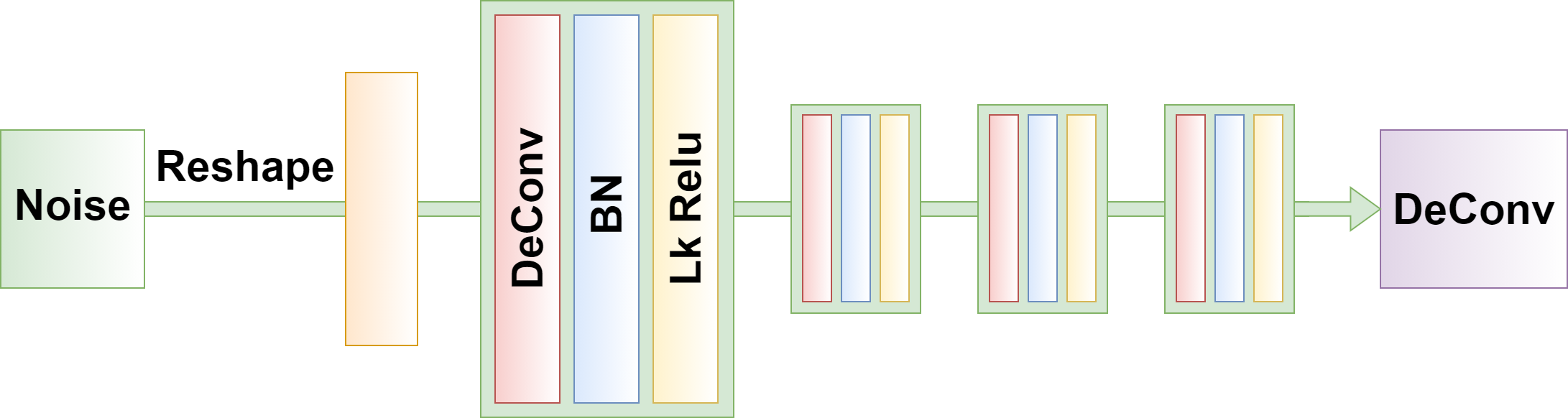}
     \caption{Our generator model consists of four deconvolution blocks, each deconvolution layer with 3 × 3 kernels.}    
  \end{figure}
After getting the trained generator, we employ the generator to produce fake images (mask images) and input them with real images into our discriminator together. 
As a result, we get two outputs of the model. $D_1(x)$ denotes the first output of the model, and $D_2(x)$ is the second output. The label of the image is denoted as y, and Gbest is the fixed generator.

The loss in step 3 is
\begin{equation}
    \begin{split}
  &L_{supervised}=-\mathbb{E}_{x,y\sim{{p}_{data}(x,y)}}\log {D}_{1}(x) \\  
  &L_{unsupervised}=\mathbb{E}_{x\sim{p}_{x}}({D}_{2}(x)-b)^{2} \\ 
  &+\mathbb{E}_{z\sim{p}_{z}}({D}_{2}({G}_{best}(z))-a)^{2} 
    \end{split}
      \label{eq:7}
  \end{equation}
The previous training stage of the generative adversarial network also played a pre-training role in the current classification.

\subsection{Pseudo-labeling}
Pseudo-labeling technology contributes a lot to the development of weakly-supervised classification. Pseudo-label learning aims at using a trained model on labeled images to predict unlabeled images. 
It filters unlabeled images based on the prediction results and feeds them into the model again for training.

We start to implement twice data augments on the fake images, then put them into our model for predicting unlabeled images. 
After summing these outputs computed by our model, we average and sharpen them to provide pseudo labels for generated images.

Ultimately, we obtain the total input images by merging the real and generated images. Correspondingly, the real and pseudo labels are used together as the total labels. 
We sum the residuals of the total input images and the scrambled total input images. Residuals of the ordered and unordered labels are summed as the total labels. 
The classification result $D_1(x)$ of the first third of the total input images x is put into the model and the first third of the total labels p is used to calculate the cross-entropy loss. 
We exploit both the classification result $D_1(u)$ of the remaining input images u obtained by the model and the remaining labels q to calculate the mean square error loss. 
With the above losses, the real images with labels and the generated images with pseudo labels can be used to train our classifier.

The loss function in Step 4 is written as:
\begin{equation}
	\begin{split}
	&L_{x}=\frac{1}{N}\sum\limits_{x\sim X}p\log{D}_{1}(x) \\ 
	&L_{u}=\frac{1}{2N}\sum\limits_{u\sim U}{\left\| q-{D}_{1}(u) \right\|}_{2}^{2} \\ 
	&loss_{4}=L_{x}+\lambda_{u}L_{u} 
	\end{split}
	\label{eq:8}
	\end{equation}
According to the above analysis, our proposed WSGAN can increase the number of medical images by generating fake images. 
It exploits contrastive learning and self-attention mechanisms to better process grayscale images with simple structure and to find lesion regions precisely. 
Moreover, the pseudo-labeling technique is applied to add labels of all unlabeled images. 
Thus, WSGAN effectively solves the three problems we mentioned in Introduction and successfully achieves weakly-supervised classification in the scenario of extreme lack of unlabeled images.

\section{Experiments}

\subsection{Medical Datasets}

Our experimental dataset includes the OCT dataset for ophthalmology, the recently more popular new Corona-virus dataset, the Brain tumor dataset, and the Chest X-ray dataset. 
The OCT dataset is divided into four categories: CNV, DME, DRUSEN, and NORMAL. Brain tumor dataset consists of glioma tumor, meningioma tumor, no tumor, and pituitary tumor. 
The Covid-19 dataset and Chest X-ray dataset are composed of two types: negative and positive.

\subsection{Semi-supervised Classification Experiment}
In this section, we conduct comparative experiments with other semi-supervised algorithms\cite{berthelot2019mixmatch,pham2021meta,nayeem2021defense,chen2020simple,miyato2018virtual} that have become more popular in recent years. 
First, we crop all image sizes to 64 x 64 and then use KNN as a semi-supervised algorithm baseline. In 1000 epochs training, our algorithm achieves satisfactory results.
\setlength\tabcolsep{1pt}
\begin{table}[h]
  \centering
    \begin{tabular}{ccccc}
    \toprule
          & OCT   & COVID-19 & Brain tumor & Chest X-ray \\
    \midrule
    KNN   & 22.70\% & 47.50\% & 16.80\% & 49.80\% \\
    SIMCLR & 35.00\% & 61.00\% & 62.60\% & 50.00\% \\
    UPS   & 37.50\% & 72.50\% & 46.70\% & 63.20\% \\
    VAT   & 32.10\% & 16.70\% & 41.70\% & 80.10\% \\
    MIXMATCH & 72.80\% & 78.50\% & 79.40\% & 96.60\% \\
    MPL   & 24.49\% & 49.71\% & 25.87\% & 64.13\% \\
    OURS  & 87.06\% & 89.50\% & 80.50\% & 96.80\% \\   
    \bottomrule
    \end{tabular}
    \caption{The top-1 accuracy of all algorithms on the entire dataset.}
    \label{tab:onelabel}
\end{table}

\noindent \textbf{Datasets.} On the OCT dataset and Chest X-Ray dataset, we uniformly sample each category to obtain 100 labeled images and 1000 unlabeled images for experimentation. 
In the brain tumor dataset, we use 100 labeled images as the labeled training dataset, and 480 unlabeled images as the unlabeled training dataset. 
For the COVID-19 dataset, due to the limitation of the number of images, we eventually select 200 labeled images and 200 unlabeled images for training.

\noindent \textbf{Setups.} We first pre-trained with unlabeled images on SIMCLR with a decoder to obtain a well-performing ResNet-50 feature extraction layer. 
The pre-train process is performed in 300 epochs on each dataset. 
Then we train the generator and discriminator according to the least square loss, and the good generator model is selected by us in line with the FID evaluation index through 3000 epochs. 
Throughout the training process, we use the Adam optimizer with a learning rate of 1 x $10^{-4}$. 
The fake images generated by the fixed generator are used together with the real images to fine-tune our classifier in the next 200 epochs, which accelerates the convergence of our model. 
In the last 1000 epochs, we use MixMatch to provide pseudo labels 

for the generated images and unlabeled images to jointly train our model for the purpose of comparison with semi-supervised classification experiments. Our experiments are performed on NVIDIA RTX 3070.

\noindent \textbf{Results.} \cref{tab:onelabel} shows the compared results of various semi-supervised classification methods on four medical datasets. 
We can observe that our algorithm outperforms most semi-supervised algorithms for medical image classification. 
Compared to MIXMATCH, our algorithm has only a slight improvement on the Brain tumor dataset and the Chest X-ray dataset, and achieve significant improvement on the OCT and COVID-19 datasets. 
The main reason is that we can obtain images through the generator, which allows us to have more unlabeled images for classification on small sample problems. 
Although MixMatch is inferior to our WSGAN, it outperforms other semi-supervised classification methods. Thus, MixMatch is a relatively effective approach. 
Despite SIMCLR can search for lesion areas by comparison between positive and negative cases, our dataset is a small dataset with few categories, resulting in a small number of negative case samples and thus SIMCLR did not achieve a good result. 
WSGAN gains good classification by combining the self-attentive mechanism, global-local structure, and contrastive learning to compensate for the lack of negative example samples in SIMCLR. 
Compared to natural images, medical images have fewer features for differentiation and smaller critical areas, leading to poor results of other algorithms on medical images.

\subsection{Weakly-supervised Classification Experiment}

In order to validate that our algorithm can perform well on classification with a small number of datasets. 
For the case of no unlabeled image (we assume that this is a case where there are very few unlabeled images), we compare our algorithm with some supervised algorithms\cite{chollet2017xception,he2016deep,jegou2017one}. 
Meanwhile, we test the performance of WSGAN in reducing the number of unlabeled images on the OCT dataset. 

\noindent \textbf{Datasets.} In this section, we only conduct experiments on the OCT dataset. Only 100 labeled OCT images are fed into our model for comparing with some supervised method.

\noindent \textbf{Setups.} In the 100-labeled image classification experiment, only labeled images are utilized in the training. In the end, we only make pseudo labels for the generated images and mix them with labeled images for classification.
\begin{table}[h]
	\centering
	  \begin{tabular}{cp{5cm}<{\centering}cp{5cm}<{\centering}}
	  \toprule
				  & Top1-acc \\
	  \midrule
	 RESNET-50    & 48.10\%  \\ 
	  XCEPTION    & 46.80\%  \\
	  DENSENET-201& 67.30\%  \\
	  OURS        & 80.90\%  \\
	  \bottomrule
	  \end{tabular}%
	  \caption{Top1-accuracy with only 100 labeled OCT images.}
	\label{tab:seclabel}%
  \end{table}

\noindent \textbf{Results.} \cref{tab:seclabel} establishes: compared with these methods, our method achieves excellent performance even when medical images are very scarce. 
Although RESNET50, XCEPTION, and DensNet networks can adequately retain image information, it is not satisfactory when medical images are very sparse. 
Our algorithm breaks the limitation of the number of training sets by adding mask images, pseudo labels to the training data, allowing WSGAN to accomplish good classification even with 100 medical images.

\begin{table}[htbp]\footnotesize
	\centering
	\caption{Add caption}
	  \begin{tabular}{cp{1cm}<{\centering}cp{1cm}<{\centering}cp{1cm}<{\centering}cp{1cm}<{\centering}cp{1cm}<{\centering}cp{1cm}<{\centering}cp{1cm}<{\centering}cp{1cm}<{\centering}}
	  \toprule
			& 1000  & 800   & 600   & 400   & 200   & 100   & 0 \\
	  \midrule
	  MixMatch & 72.80\% & 71.80\% & 73.00\% & 69.20\% & 64.90\% & 67.90\% & - \\
	  \textbf{Ours} & \textbf{87.06\%} & \textbf{85.90\%} & \textbf{83.30\%} & \textbf{83.19\%} & \textbf{82.30\%} & \textbf{81.50\%} & \textbf{80.90\%} \\
	  \bottomrule
	  \end{tabular}
	\label{tab:addlabel}
  \end{table}
From \cref{tab:addlabel}, we can obtain the following observations:

(1)	When we have 1000 unlabeled OCT images, the accuracies of our algorithm and MIXMATCH are 87.06\% and 72.80\%, respectively. 
This means that when there are enough unlabeled images, WSGAN shows better performance than MIXMATCH by generating semantically rich feature images

(2)	The accuracy of both MIXMATCH and WSGAN show a decreasing trend as the number of unlabeled images decreases. 
Such a decreasing trend, however, is different. Despite WSGAN showing a smooth decline and MIXMATCH showing a fluctuating decline, the accuracy of our algorithm is always higher than that of MIXMATCH. 
This seems to indicate that contrastive learning and self-attention mechanism also play an important role in classifying the medical images.

(3)	When we do not use unlabeled images, WSGAN can still maintain a high accuracy due to the sufficient fake images and pseudo labels. 
However, it is ineffective for MIXMATCH to deal with the scenario of lacking unlabeled images. Generally, it is widespread that we always haven’t sufficient unlabeled images in medical applications. 
Therefore, our algorithm has a wider and more effective application.

\subsection{Ablation Experiment}
In this section, we perform two main tasks. The first one is that we discuss the correctness of using trained generators to provide images for the classifier instead of classifying them at the time of generation; 
the second one is that we analyze the impact of adding decoders to self-supervised learning on solving the problem of pattern collapse.

\begin{figure}[h]
  \centering
   \includegraphics[width=0.7\linewidth]{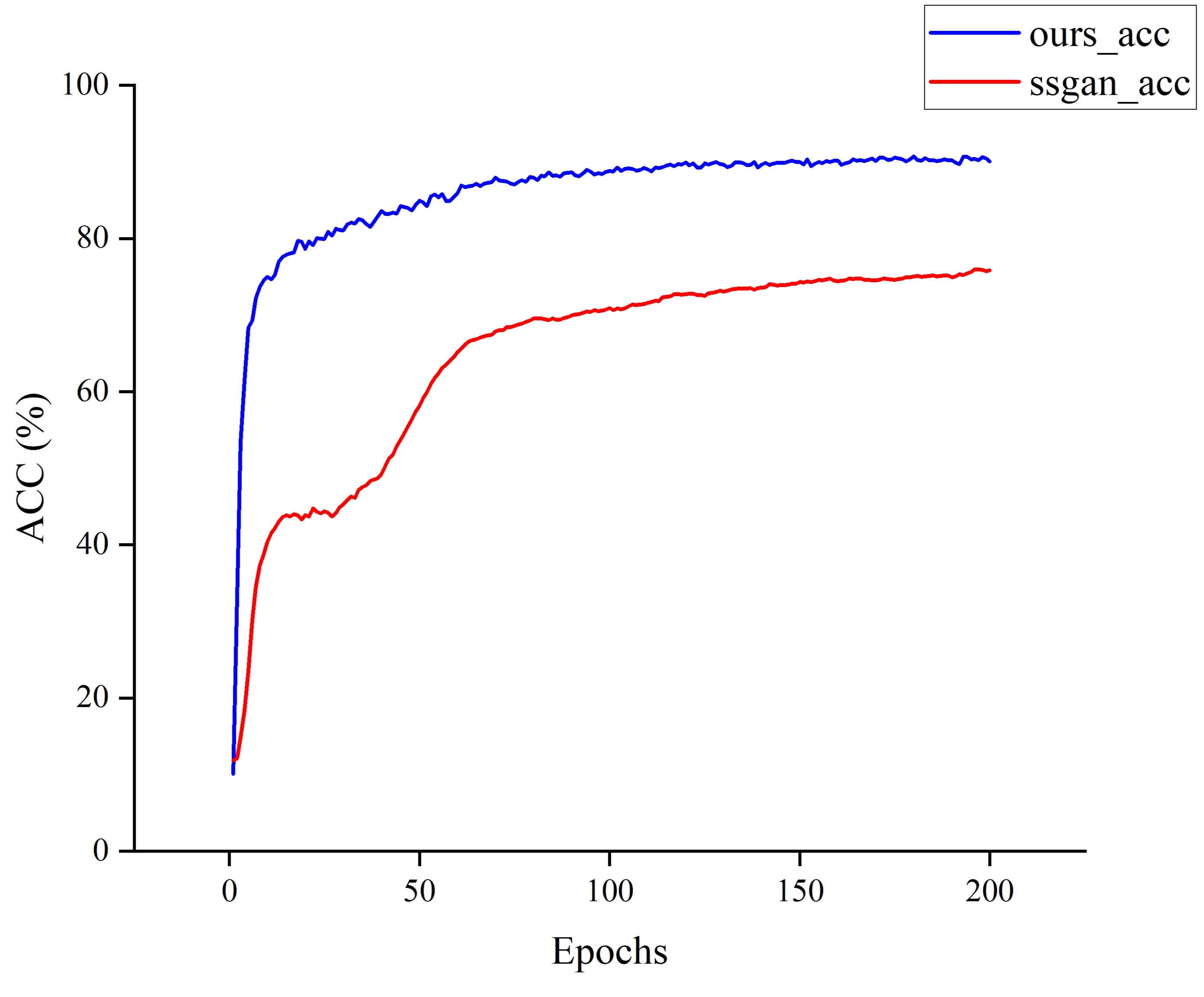}
  \caption{MNIST accuracy.}
   \label{figure:tencol}
\end{figure}

\noindent \textbf{(1)	Discriminator classifies the generated images.} First, we discuss the generator and discriminator of SSGAN\cite{odena2016semi}. 
SSGAN performed image generation and classification simultaneously. Its generator is also be concerned with the accuracy of classification when performing image generation, 
while the classifier is used to measure the quality of the generated images when performing image classification. In the end, the training can only get a poor generator and classifier. 
As we can see later, SSGAN provides poor learning performance with 1000 labeled images and 1000 unlabeled images on MNIST and CIFAR-10. 
We only exploit generative adversarial loss and FID evaluation metrics to obtain better performing generator by training 1000 epochs. 
During next 1000 epochs, the trained generator is fixed, and we train the classifier with a semi-supervised classification generative adversarial loss. It is seen from \cref{figure:tencol} that the learning performance of SSGAN is enhanced. 

\begin{figure}[h]
	\centering
	 \includegraphics[width=0.7\linewidth]{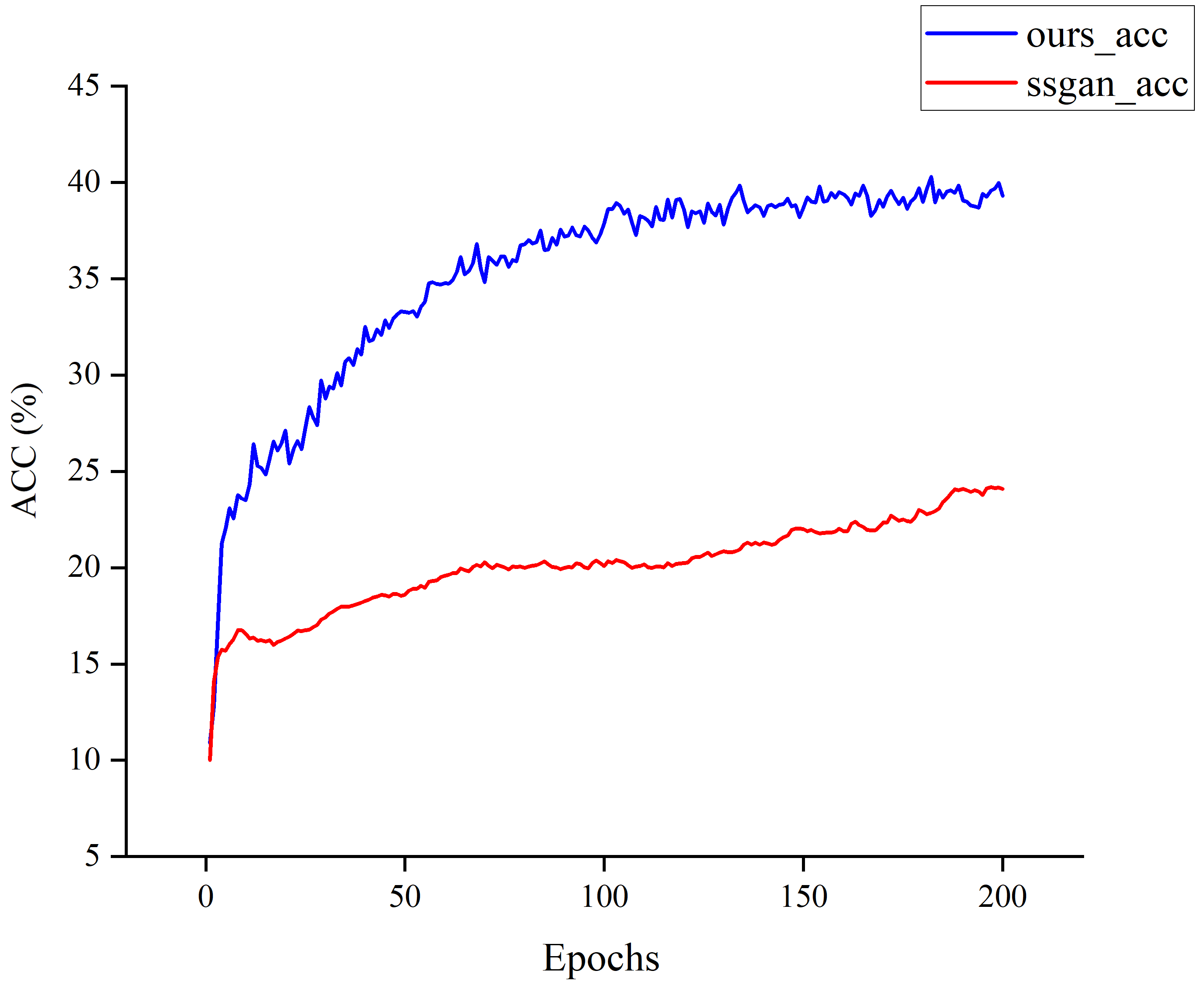}
	\caption{CIFAR-10 accuracy.}
	 \label{figure:ninecol}
  \end{figure}
\noindent \textbf{(2)	Contrastive Learning with a Decoder.} We test the effect of adding decoders on improving the model to solve the mode collapse issue by comparing the standard deviation (std) of the last layer of Z and contrastive loss during contrastive learning pre-training.
 We perform 100 and 500 training epochs on the OCT dataset and the chest X-ray dataset, respectively, compare SIMCLR with and without decoders.
\begin{figure}[h]
	\centering
	 \includegraphics[width=0.7\linewidth]{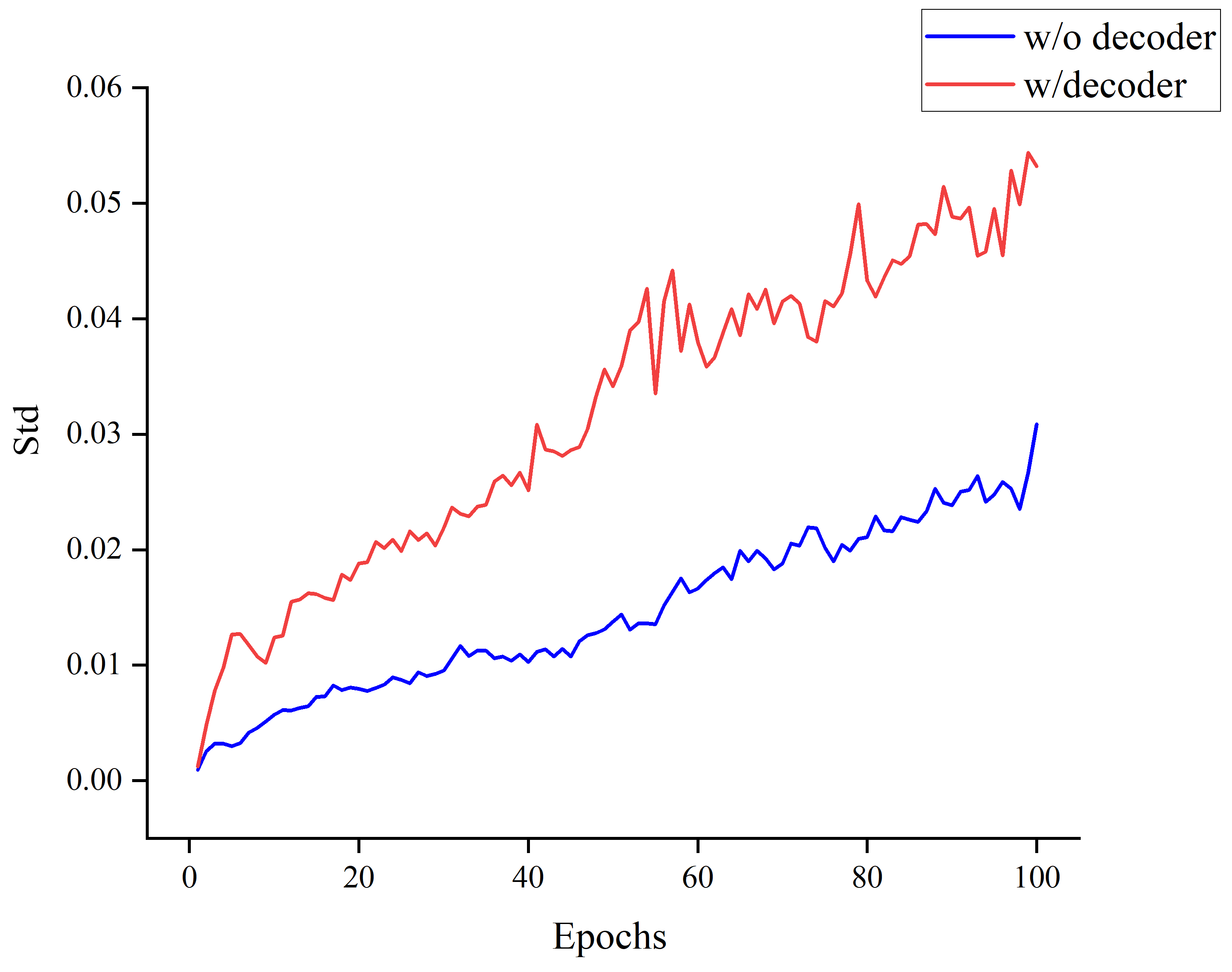}
	\caption{The std of Z on the OCT dataset.}
	 \label{figure:12col}
  \end{figure}
  \begin{figure}[h]
	\centering
	 \includegraphics[width=0.7\linewidth]{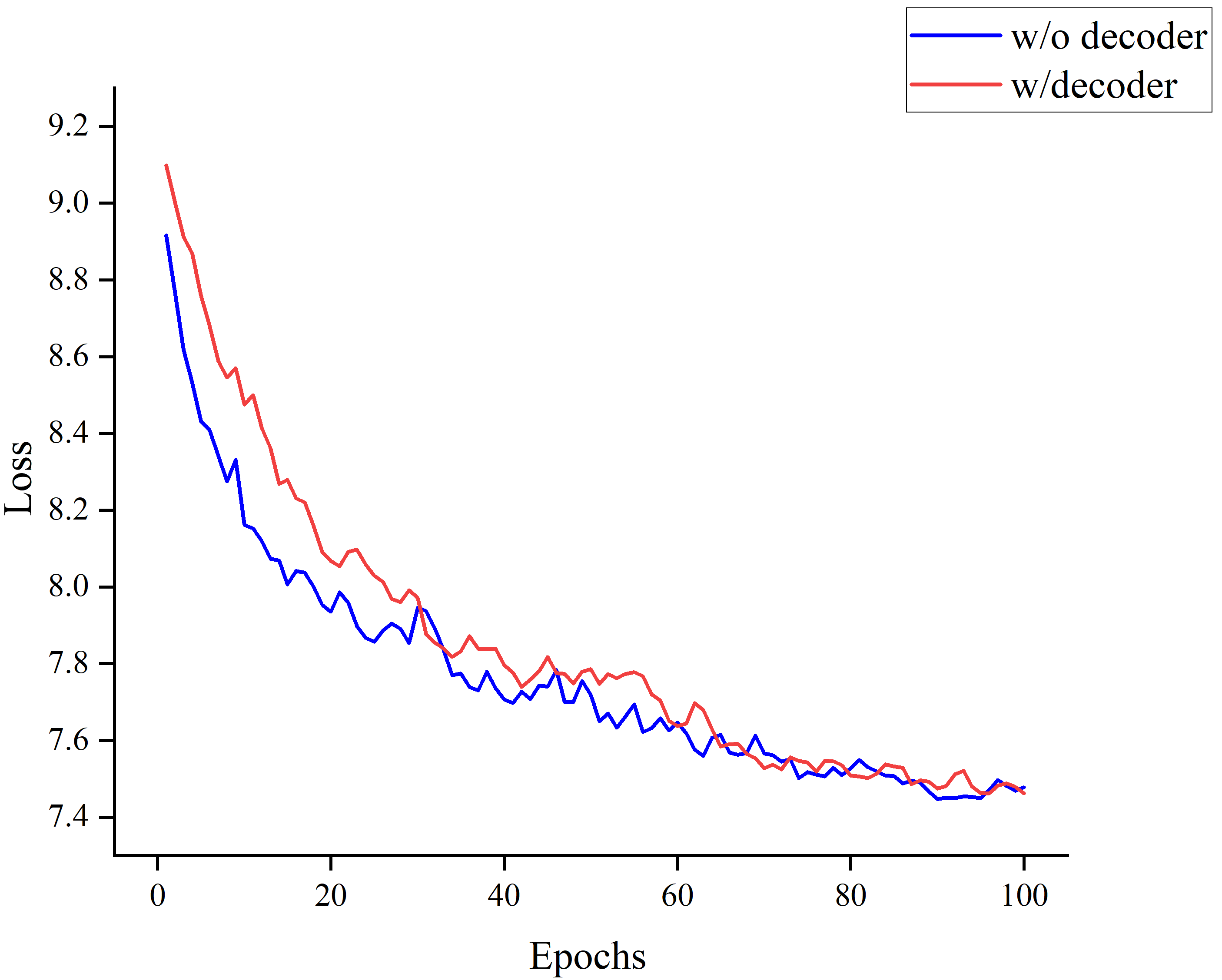}
	\caption{The contrastive loss on the OCT dataset.}
	 \label{figure:13col}
  \end{figure}
  \begin{figure}[h]
	\centering
	 \includegraphics[width=0.7\linewidth]{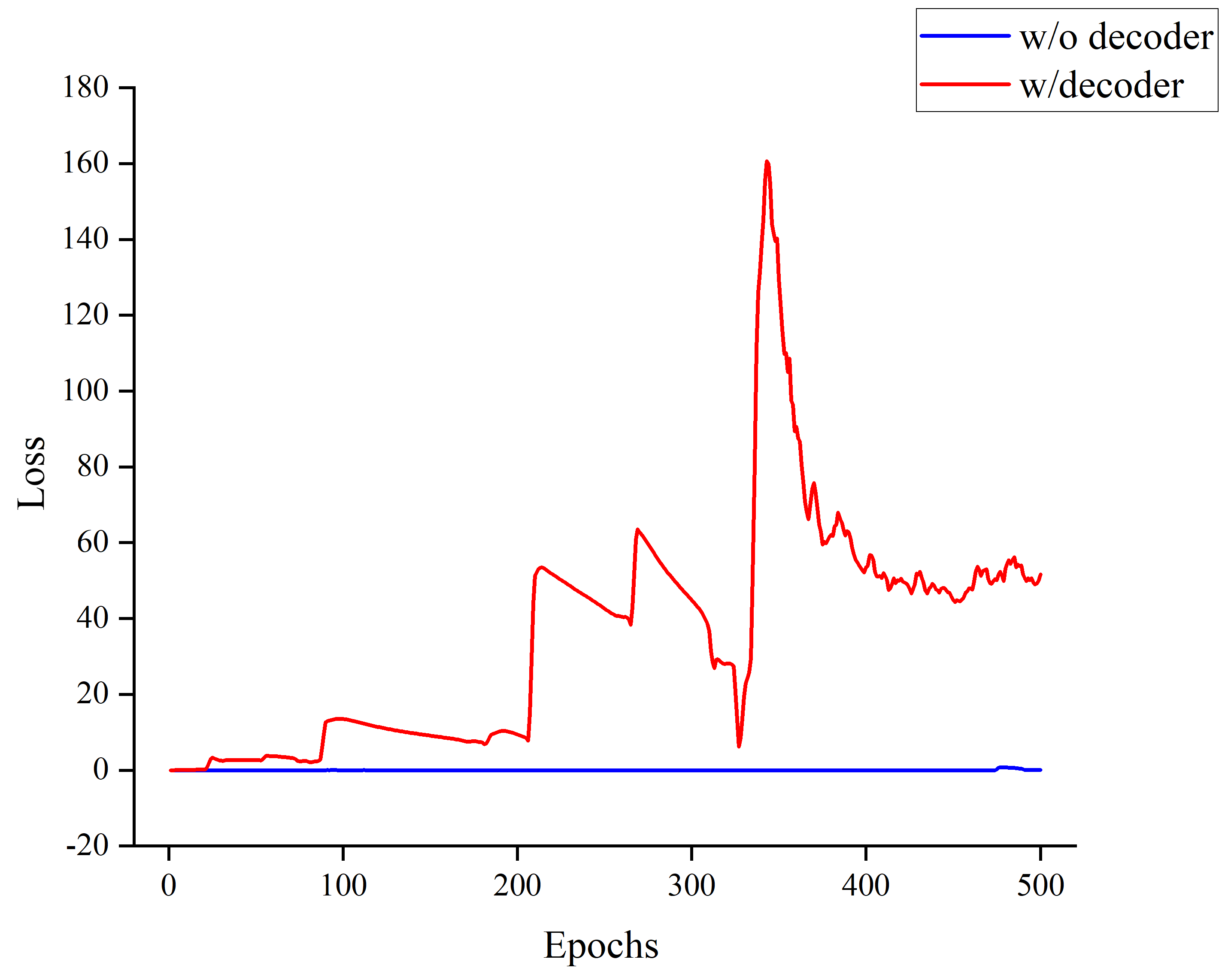}
	\caption{The std of Z on the Chest X-ray dataset.}
	 \label{figure:14col}
  \end{figure}
  \begin{figure}[h]
	\centering
	 \includegraphics[width=0.7\linewidth]{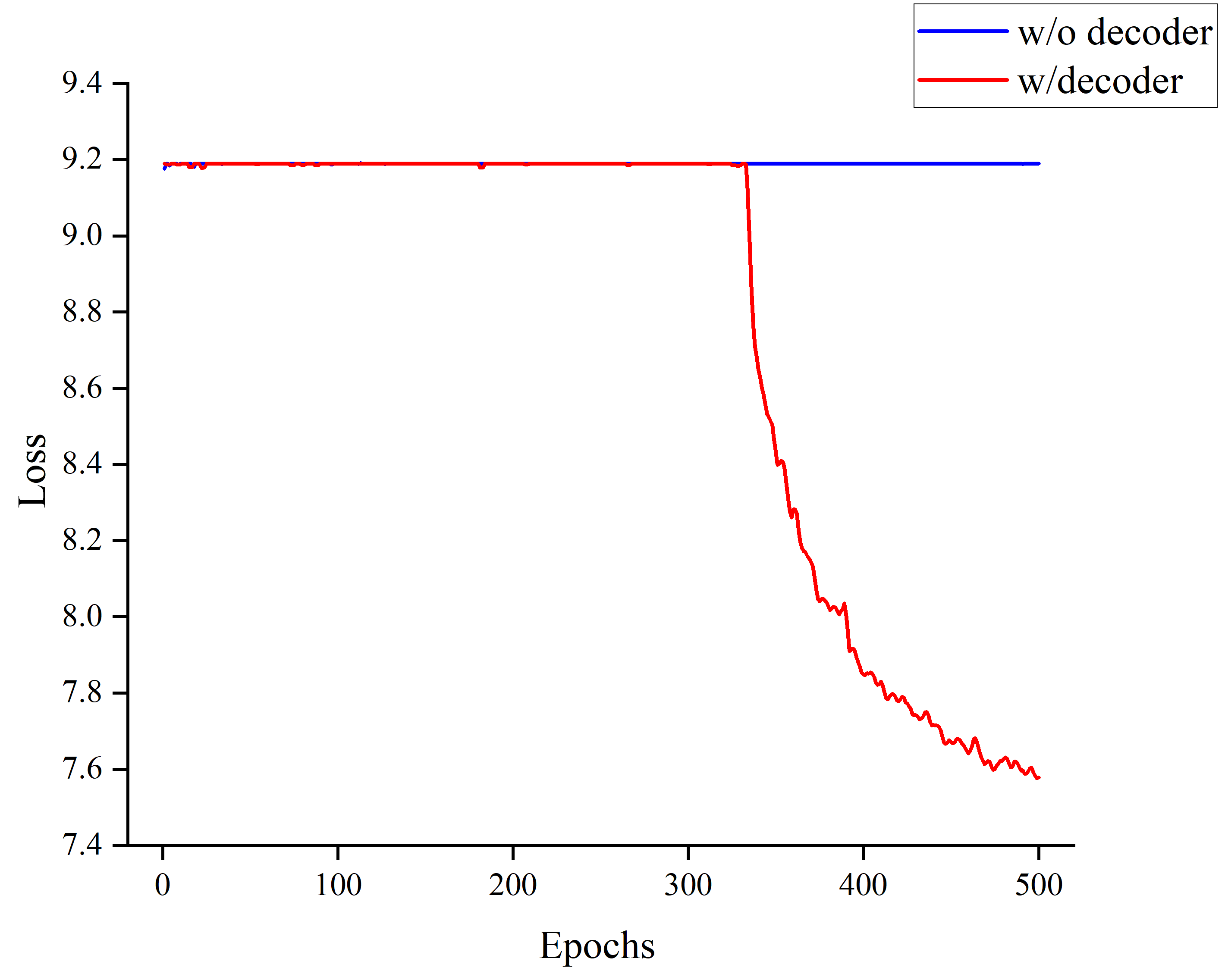}
	\caption{The contrastive loss on the Chest X-ray dataset.}
	 \label{figure:15col}
  \end{figure}

According to \cref{figure:12col} and \cref{figure:13col}, on the brain tumor dataset, SIMCLR has a certain ability to solve the problem of pattern collapse due to the variety of brain tumor datasets, 
which increases the number of negative samples. After adding a decoder, it can improve SIMCLR's ability to solve the mode crash problem.

\begin{figure}[h]
	\centering
	 \includegraphics[width=1\linewidth]{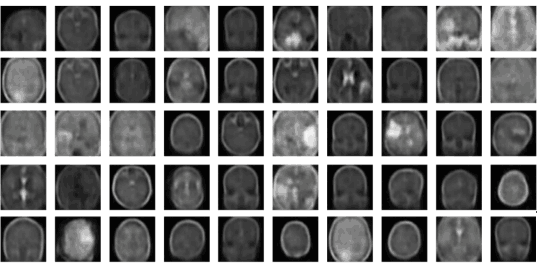}
	 \caption{Decode images of middle layer H.}
	 \label{figure:16col}
  \end{figure}

In \cref{figure:14col} and \cref{figure:15col}, because the chest X-ray dataset has only positive and negative points, thus there are fewer negative cases, which increases the odds of SIMCLR model collapse. 
After the decoder is added, due to the limitation of cyclic consistency loss, the output of the last layer cannot be the same constant. 
Therefore, contrastive learning with the decoder can boost the model's ability to solve the mode collapse. 
Following the output of the decoder in \cref{figure:16col}, we can see that the middle layer H can well represent the effective content of the image.
\section{Conclusion}
Aiming at the problem of medical data with fewer labels and images, we propose to combine the popular adversarial learning with weakly-supervised learning, 
so that the GAN is capable of generating sufficient images for scarce medical datasets. 
At the same time, pseudo-labeling technology is utilized to provide labels for the generated images to achieve weakly-supervised classification. 
We find the key regions in medical images through contrast learning and self-attention mechanism, 
which successfully addresses the problems of few identifiable features and small lesion areas in medical images to a certain extent. 
Extensive experimental results on four medical image datasets demonstrate the effectiveness of our approach. 
In the future, we will consider how to improve learning performance by generating high-definition images with only a few images.

\section*{Acknowledgement}
This work was supported by Public Projects of Zhejiang Province under Grant GG22F025981.

{\small
\bibliographystyle{ieee_fullname}
\bibliography{egbib}
}

\end{document}